\documentclass[12pt]{article}\usepackage{authblk}

\usepackage{times}
\usepackage{graphicx}
\usepackage{amsmath}
\usepackage{amssymb}
\usepackage{cite}
\usepackage{url}
\usepackage{setspace}
\usepackage{lineno}
\usepackage{natbib}
\usepackage{authblk}
\usepackage{comment}
\usepackage{multirow}

\usepackage{titlesec}

\titlespacing*{\paragraph}{0pt}{0pt}{4pt}

\usepackage[table]{xcolor}
\usepackage{colortbl}
\usepackage{color}

\usepackage{changepage}
\usepackage{makecell}
\usepackage{mathptmx}
\usepackage{comment}
\usepackage{url}
\usepackage{hyperref}

\usepackage{multirow}
\usepackage{xcolor}
\hypersetup{
    colorlinks,
    linkcolor={black!50!black},
    citecolor={blue!50!black},
    urlcolor={blue!50!black}
}

\definecolor{lightgrey}{RGB}{245,245,245}
\definecolor{white}{RGB}{255,255,255}

\title{Modeling Public Perceptions of Science in Media}
\date{}

\author[1]{Jiaxin Pei}
\author[2]{Dustin Wright}
\author[2]{Isabelle Augenstein}
\author[3]{David Jurgens}
\affil[1]{Stanford University}
\affil[2]{University of Copenhagen}
\affil[3]{University of Michigan}

\begin{document}

\maketitle
\vspace{-1.5cm}
\begin{center}
\texttt{pedropei@stanford.edu, dw@di.ku.dk, augenstein@di.ku.dk, jurgens@umich.edu}
\end{center}

\begin{abstract}
    Effectively engaging the public with science is vital for fostering trust and understanding in our scientific community. Yet, with an ever-growing volume of information, science communicators struggle to anticipate how audiences will perceive and interact with scientific news. In this paper, we introduce a computational framework that models public perception across twelve dimensions, such as newsworthiness, understandability, importance, and surprisingness. Using this framework, we create a large-scale science news perception dataset with 10,489 annotations from 2,101 participants who rated 1,506 science news stories. This dataset, sourced from diverse US and UK populations, provides valuable insights into public responses to scientific information across multiple domains. We further develop natural language processing models that predict public perception scores with a strong performance. 
    Leveraging the dataset and model, we examine public perception of science from two perspectives: (1) Perception as an \textit{outcome}: What factors affect the public perception of scientific information? (2) Perception as a \textit{predictor}: Can we use the estimated perceptions to predict public engagement with science? We find that individuals’ frequency of science news consumption is the strongest driver of perception, whereas demographic factors exert minimal influence. More importantly, through a large-scale analysis and carefully designed natural experiment on Reddit, we demonstrate that the estimated public perception of scientific information has direct connections with the final engagement pattern. Posts with more positive perception scores receive significantly more comments and upvotes, which is consistent across different scientific information and for the same science, but are framed differently. 
    Overall, this research underscores the importance of nuanced perception modeling in science communication, offering new pathways to predict public interest and engagement with scientific content.

    \end{abstract}

\section{Introduction}
Science communication plays a key role in disseminating scientific information and promoting public engagement with science \citep{fischhoff2013sciences}. Effectively communicating science to the public holds significance for both scientists and the broader audience outside the science community. On the one hand, individuals living in modern society rely on scientific knowledge to make important life decisions \citep{bell2003understandings}, and in general, the public shows great interest in scientific information \citep{durant1989public}. On the other hand, public engagement and discussions about scientific information could also help to promote societal understanding of science and inform new scientific developments \citep{stilgoe2014should}.

Despite its crucial role in our society and the considerable efforts devoted to it, effectively communicating science to the public remains challenging and encounters numerous obstacles \citep{cheng2008communicating, scheufele2013communicating}. One major challenge for science communicators is understanding the public \citep{schafer2018different}. Assumptions about the public represent an important latent dimension that distinguishes between the two classic models of science communication. In the deficit model, the public is simply viewed as a group of individuals that are passive and ignorant, and science communication is to simply ``sell science'' to the public \citep{sturgis2004science}. Such a model has long been criticized because it overlooks the diversity of the audience \citep{sturgis2004science}. The engagement model calls for science communicators to better contextualize the communication of science to promote science engagement with diverse groups of people \citep{stilgoe2014should}. However, in practice, understanding the public is challenging, even for professional journalists \citep{atkin1983journalists}. Additionally, existing research also suggests that many current science communication practices may only attract people who are already well-informed about certain science issues but fail to engage with a broader and more diverse set of audiences \citep{goidel2006exploring, nisbet2009s}. Therefore, understanding the broad audience of science communication could provide further insights into facilitating effective science communication.

Existing studies have explored the interactions between science and diverse audiences from two primary perspectives. Firstly, researchers have investigated how various social factors influence people's attitudes and attention toward scientific information \citep{weinburgh1995gender, achterberg2017science, mccright2013influence}. These studies often employ surveys, where participants rate specific scientific issues such as gene technology \citep{olofsson2006attitudes} and climate change \citep{shi2015public}. Another relevant area of research examines how different framing and language usage impact people's responses to scientific information \citep{feldman2021upping}.
However, within this line of research, while some studies have explored how pre-existing beliefs or political orientations shape individuals' consumption of scientific information \citep{yeo2015selecting, hart2012boomerang}, there remains a gap in understanding the nuanced interactions between content and individual backgrounds. In short, we identify three major gaps in existing studies: (i) a tendency to narrowly focus on very specific topics (e.g. climate change), overlooking the nuanced variations across and within domains; (ii) the nuanced interactions between people's backgrounds and individual scientific information presented in media are underexplored; (iii) a lack of datasets and computational tools to understand public perception of scientific information in media at scale. In this study, we aim to help promote public engagement with science by addressing two questions:  (1) What factors affect the public's perception of scientific information in the media? (2) Do scientific information with more positive perceptions receive more public engagement?

We introduce a computational pipeline designed to assess the public perception of scientific information in the media. Firstly, we propose a novel comprehensive framework comprising 12 dimensions: \textsc{Newsworthiness}, \textsc{Understandability}, \textsc{Expertise}, \textsc{Interestingness}, \textsc{Fun}, \textsc{Importance}, \textsc{Benefit}, \textsc{Sharing}, \textsc{Reading Willingness}, \textsc{Exaggeration}, \textsc{Surprisingness}, and \textsc{Controversy}. Following this framework, we curate a large-scale science news perception dataset containing 10,489 annotations from 2,101 participants across 1,506 science news stories. Our dataset draws from representative samples of the US and UK populations, ensuring diversity and representativeness in measuring public perceptions. Each participant evaluated a variety of science news stories based on our proposed framework. Finally, we develop NLP models to automatically predict public perception of science news. 

This computational framework allows us to systematically examine the public perception of science from two perspectives: (1) Perception as an outcome: What factors affect people's perception of scientific information in media? (2) Perception as a predictor: Can estimated perceptions predict public engagement with scientific information?
We find that individuals' science news consumption frequency is the strongest factor affecting their perceptions of science news stories, while demographic factors only have a minimal effect. Furthermore, through a large-scale analysis and a natural experiment on Reddit, we examine the connection between the estimated perceptions and the engagement patterns with scientific information. We find that the potential perceptions of a post correlate with public engagement with it. For example, posts with higher estimated \textsc{Surprisingness}, \textsc{importance}, and \textsc{Fun} tend to receive more upvotes and comments on Reddit. Such a pattern persists even when we compare different framings of the same scientific information, suggesting a strong causal relationship between the perception and engagement.  Overall, our analysis highlights the significance of modeling public perception of scientific information in the media. It not only helps to better understand the public interest in science but also holds the potential to predict engagement patterns with scientific information.

\section{A Multi-dimensional Framework for Modeling the Public Perception of Scientific Information in Media}
\label{sec:framework_news_value}

Modeling public perception of scientific information is a key step towards effective science communication \citep{longnecker2023good}. However, public perception is a subjective and multi-dimensional construct. Here, drawing on existing literature on news values, public understanding of science, and science communication, we develop a comprehensive framework to analyze the public perception of scientific information in the media.

\paragraph{\textsc{Newsworthiness}} In science journalism practices, newsworthiness is regarded as a core dimension that guides the selection and coverage of information \citep{harcup2017news}. Similarly, scholars have studied the news values of scientific information in news stories \citep{badenschier2011issue} or research articles \citep{nishal2022crowd}. While many existing studies focus on modeling journalists' understanding of newsworthiness, some also emphasize the importance of studying perceived newsworthiness in various settings \citep{lee2014newsworthy}.
Here, we approach the newsworthiness of scientific information from the perspective of the general public. Specifically, our goal is to measure whether the public views a news article as ``should be published in news or not''. \textsc{Newsworthiness} represents the holistic evaluation of the perceived value of the presented information.

\paragraph{\textsc{Understandability}}  In science communication, presenting content that is accessible and comprehensible to the audience is crucial, as the value of scientific information in news is contingent on its understandability \citep{fischhoff2012communicating}. 
Without clear comprehension, the lay public is unlikely to benefit from the scientific news presented and further participate in the public discussion about it \citep{varner2014scientific}. However, despite existing efforts to make science more accessible to the public, existing studies suggest that understandability remains a challenge in current science news reporting \citep{morosoli2024scientific}. Therefore, it is important for us to measure how the public perceives the understandability of science presented in the media.

\paragraph{\textsc{Expertise}} Scientific information usually contains jargon or specialized knowledge that the general public may not easily understand \citep{hirst2003scientific, sharon2014measuring}. Here, we aim to quantify whether a given science news story requires specialized knowledge for the public to understand. \textsc{Understandability} models an individuals' capability to understand the content, which could also depend on the individual's background in the relevant topic. Unlike understandability, \textsc{expertise} focuses on the relatively more objective ``difficulty'' of the content.

\paragraph{\textsc{Importance}}  While scientific findings may be highly valued within specific scientific communities, their significance to the broader society can vary. Thus, assessing whether the science covered is of sufficient importance to the general audience is crucial in understanding the overall perception of science news \citep{burns1995newspaper}. Importance has been considered as a key dimension in evaluating science news values \citep{badenschier2011issue} and here we focus on the public perception of scientific importance. By doing so, we aim to gain insights into how the public perceives the significance of research outputs presented in news media.

\paragraph{\textsc{Fun}} Entertainment is recognized as an essential news factor in journalism \citep{harcup2017news}. Despite science often being seen as a serious topic, making scientific information engaging is essential for science communicators to foster public engagement with science \citep{stilgoe2014should, fischhoff2013sciences}. Therefore, assessing the entertainment value or the fun aspect of reading science news stories is important in evaluating the overall perceptions.

\paragraph{\textsc{Surpringness}}
Surprisingness, characterized by content that contrasts with people's existing knowledge or perceptions, is widely recognized as a core dimension of news value \citep{galtung1965structure}. Here, we aim to measure how the public perceives the surprisingness of scientific information presented in media. An existing study suggests that findings that surprise the audience may be deemed to have a higher news value \citep{htoo2023medical}, as they not only inform but also engage by introducing novel perspectives or unexpected information.

\paragraph{\textsc{Controversy}} Scientific information in the media can be controversial due to the natural uncertainty of scientific research and the significance of the topic it tackles \citep{sarewitz2004science, friedman1999communicating}. Furthermore, existing studies also suggest that controversies discussed in the media can also potentially affect scientific publications \citep{stewart2008media}. Therefore, it is important to measure how the public perceives the scientific information presented in news media to be controversial.

\paragraph{\textsc{Exaggeration}} Science reporting has long grappled with the issue of exaggeration or hype \citep{caulfield2012science}. While exaggeration may initially attract more audience attention, existing studies suggest that perceived exaggeration could erode public trust in science \citep{intemann2022understanding, neureiter2021trust}. Hence, measuring how the public perceives exaggerations in science news stories is crucial for a comprehensive evaluation of scientific information in the media.

\paragraph{\textsc{Interestingness}} \textsc{Interestingness} represents a subjective but essential dimension in assessing science news, tailored to the preferences of specific audiences. In journalism research, perceived interestingness is considered a core factor affecting people's perceptions of overall newsworthiness \citep{lee2014newsworthy}. One of the primary goals of science communication is to engage the public with information that captures their interest; therefore, it is essential for us to measure the perceived interestingness of scientific information presented in news articles.

\paragraph{\textsc{Benefit}} Scientific knowledge plays a key role in informing individuals' decision-making and facilitating societal transformations \citep{yin2022public}. Consequently, the direct applicability or usability of science news stories becomes a critical dimension of their value \citep{badenschier2011issue}. This dimension underscores the importance of presenting scientific findings in a way that is not only informative but also practically relevant to the audience's daily lives and decisions.

\paragraph{\textsc{Sharing Willingness}} 
Social media platforms have gradually become one of the primary sources of news information for the general public \citep{shearer2019americans}. Therefore, studying the sharability of news stories and users' sharing behaviors has attracted attention from journalism scholars \citep{bednarek2016investigating, harcup2017news}.  In this context, we consider the likelihood of readers sharing scientific information as an important dimension to measure the overall perception of science news, reflecting the impact and reach of science communication in the digital age. We use \textsc{Sharing} to represent this category in the rest of the paper.  

\paragraph{\textsc{Reading Willingness}} Reading represents one of the most basic types of engagement with scientific information. \textsc{Reading Willingness} measures whether the public would potentially be willing to read about the content in different venues. Good science news should attract the audience and prompt them to read further about the topic. Articles that the audience does not even want to read are less likely to facilitate further discussions and engagements with the corresponding topic. Therefore, we consider reading willingness as an important dimension in our framework. We use \textsc{Reading} to represent this category in the rest of the paper.

Collectively, the 12 dimensions provide a comprehensive evaluation of public perceptions of scientific information and can be used to measure science presented in diverse types of media, including news stories, social media posts, and research articles.  Appendix \ref{supp:framework-table} details the descriptions of each dimension and the statements used to measure them.

\section{Results}
How the public perceives scientific information has long been a core question for the science community \citep{national2017communicating}. In this study, we conduct a series of experiments to investigate how factors like demographics and trust in science affect public perception of scientific information and the connection between perception and downstream engagement on social media.  More specifically, we focus on two research questions: \textbf{RQ1 Perception as an outcome:} What factor has the largest influence on people's perception of science news? \textbf{RQ2 Perception as a predictor:} Are there any connections between the estimated perceptions and the final public engagement with scientific information?

\subsection{Analyzing Factors Affecting the Public Perceptions of Science}
\begin{figure*}[h]
 \centering
  \includegraphics[width=\textwidth]{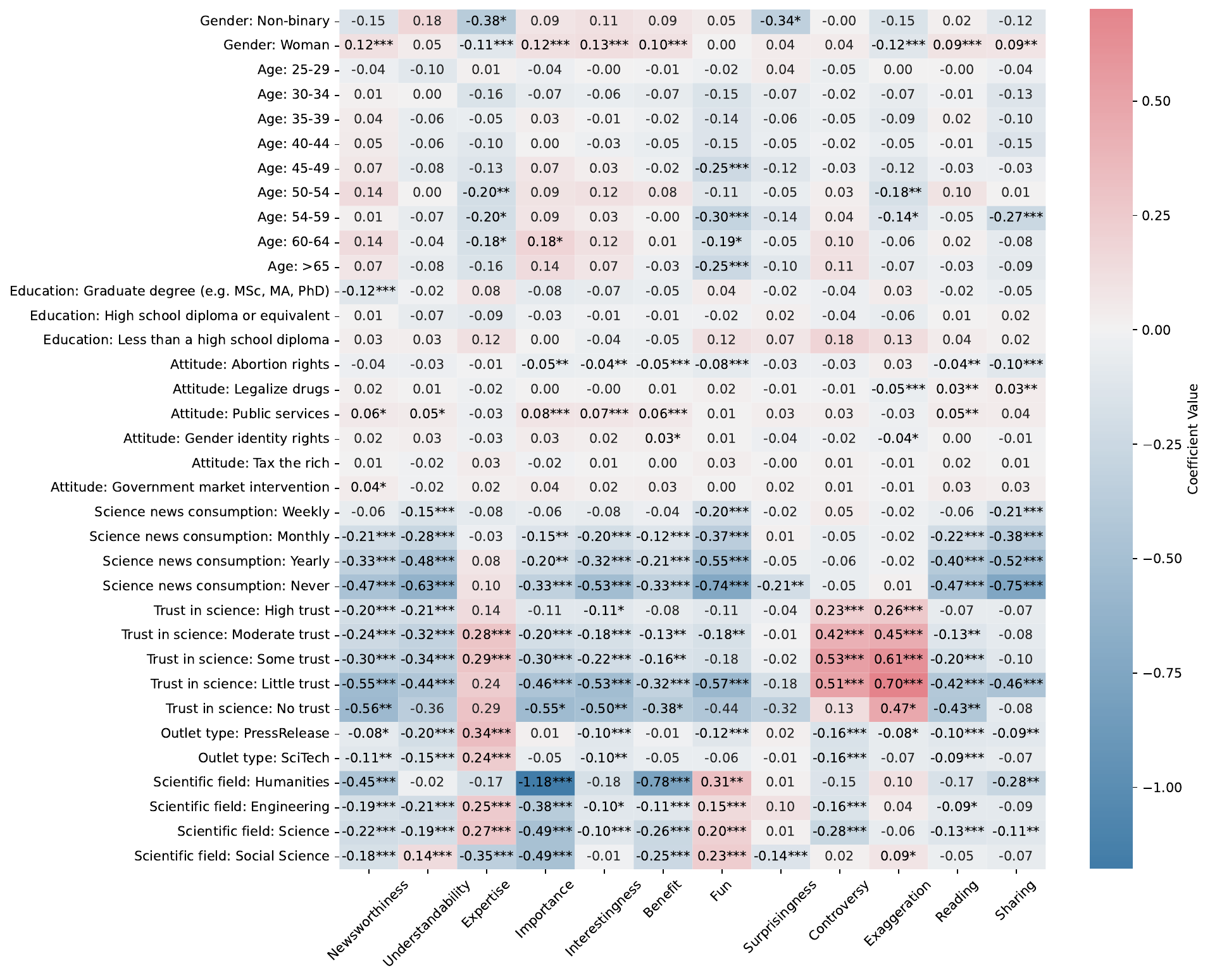}
 \caption{Experience with science and individuals' backgrounds could both affect people's perception of scientific information. People's science news consumption frequency and trust in science have the largest effect sizes. }
\label{fig:reg_content_user_factor_newsvalues}
\end{figure*}

Public attitudes toward science are subjective and can be affected by individuals' backgrounds such as gender \citep{weinburgh1995gender} and education \citep{bak2001education}. However, existing studies mainly focus on people's broad attitudes toward science, while science comes with large differences across fields and communities \citep{yin2022public}.
Therefore, it is important to study how different background factors affect people's perceptions of scientific information presented in media. In this study, we examine the connection between background factors and public perceptions of scientific information in news articles using the nuanced background information as well as individual ratings for science news stories. 
To understand how background factors affect people's perceptions of individual science news stories, we construct separate mixed-effect regression models leveraging the background factors to predict each dimension of public perceptions. The background factors include three major dimensions: (1) demographic information, including gender, age, and educational level; (2) experience with science, including science news consumption frequency and trust in science; (3) political orientations, including the attitudes towards women's reproductive health rights, legalizing drugs, government investments in public services, gender identity rights, taxing the rich and government interfering with the market.  To account for the effect of article-level variations, we control the article ID as the random effect. Additionally, we also control the scientific field and the outlet type as fixed effects. Figure \ref{fig:reg_content_user_factor_newsvalues} presents the aggregated results from the regressions.

\begin{figure*}[t]
 \centering
  \includegraphics[width=\textwidth]{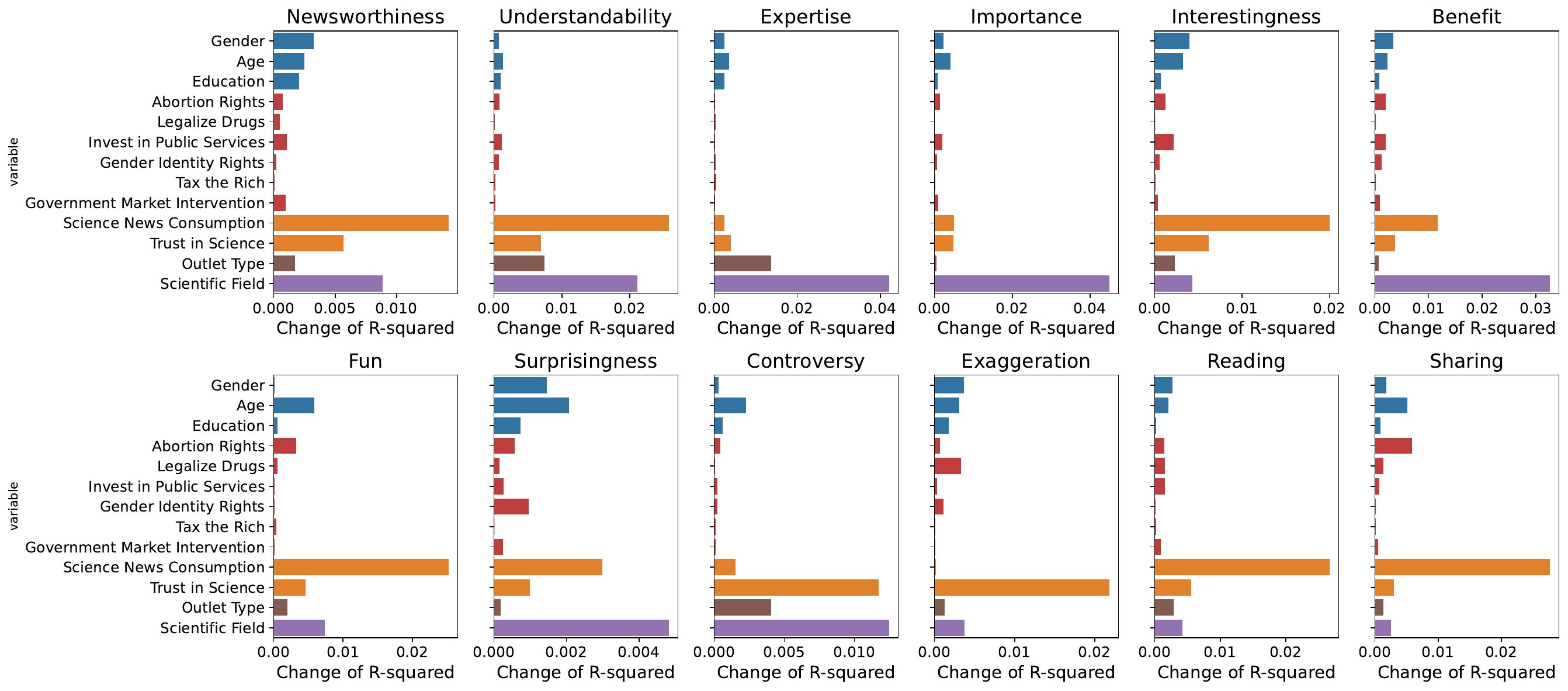}
 \caption{Individual's science news consumption frequency, trust in science, and the domain of scientific information are the strongest predictors of people's perception of scientific information in the media. Surprisingly, the widely studied demographic factors and political orientation factors have minimal effects. }
\label{fig:reg_newsvalues_change_in_r2}
\end{figure*}

\paragraph{Science News Consumption Frequency has the largest effect on people's perception of scientific information} As shown in Figure \ref{fig:reg_content_user_factor_newsvalues}, science news consumption frequency is the driver of individuals' rating for scientific information. For example, people who consume science news on a daily basis tend to rate the same scientific information 0.47 higher than people who never consume science news stories on a 5-point scale. The effect size becomes even larger for \textsc{Fun} and \textsc{Sharing}, suggesting that people's science news consumption frequency is significantly associated with their tendency to enjoy and share science. Similarly, trust in science also has a large effect on people's perceptions of scientific information regarding newsworthiness and interestingness. Additionally, trust in science significantly affects individuals' ratings for controversy and exaggeration, while science news consumption does not. This result suggests that prior trust in science could potentially affect individuals' trust in scientific information. Overall, our results suggest that individuals' science news consumption frequency and trust in science play a significant role in shaping people's perceptions of scientific information compared with other background factors.

\paragraph{Gender, age, education, and political attitudes have relatively small effects on public perception of science} 
Existing studies have widely studied how demographic factors affect people's perception of science and scientific information \citep{osborne2003attitudes, weinburgh1995gender, jones2000gender, blank2015does}. Here, we present the most comprehensive analysis on demographic factors, with the control of many other factors, including content and media type. Surprisingly, we found that age, education level and most of the political orientation factors do not significantly affect individual ratings for science news stories on most of the dimensions. Furthermore, while some existing studies suggest that men hold more positive attitudes towards science \citep{weinburgh1995gender, jones2000gender}, our results suggest that women may actually show greater interest in scientific information presented in media. Moreover, while existing studies generally suggest a positive correlation between people's education level and their attitudes towards science \citep{osborne2003attitudes}, we observe that people with a graduate degree may actually consider the same scientific information with lower newsworthiness compared with individuals with a college degree. One potential explanation is that people with a higher degree may have a more stringent perception of news values due to their graduate-level training.

\paragraph{Science domain and media type strongly affect public perception of science} In comparison to general news outlets, science news stories presented in Press Releases and SciTech blogs are consistently perceived as less newsworthy. This may be attributed to the more technical nature of the content covered by these outlets, which may be less understandable to the general public. 
Larger differences are observed in the scientific domain of the content. Although the Humanities are perceived as more fun and understandable, they are considered significantly less important and less newsworthy than other subjects. In contrast, science news stories about medicine are perceived as more important and thus considered more newsworthy compared to other subjects. The effect size of domains is significantly larger than that of other demographic information, indicating that content plays a more substantial role in influencing how the public perceives scientific information.

\begin{figure*}[t]
 \centering
  \includegraphics[width=\textwidth]{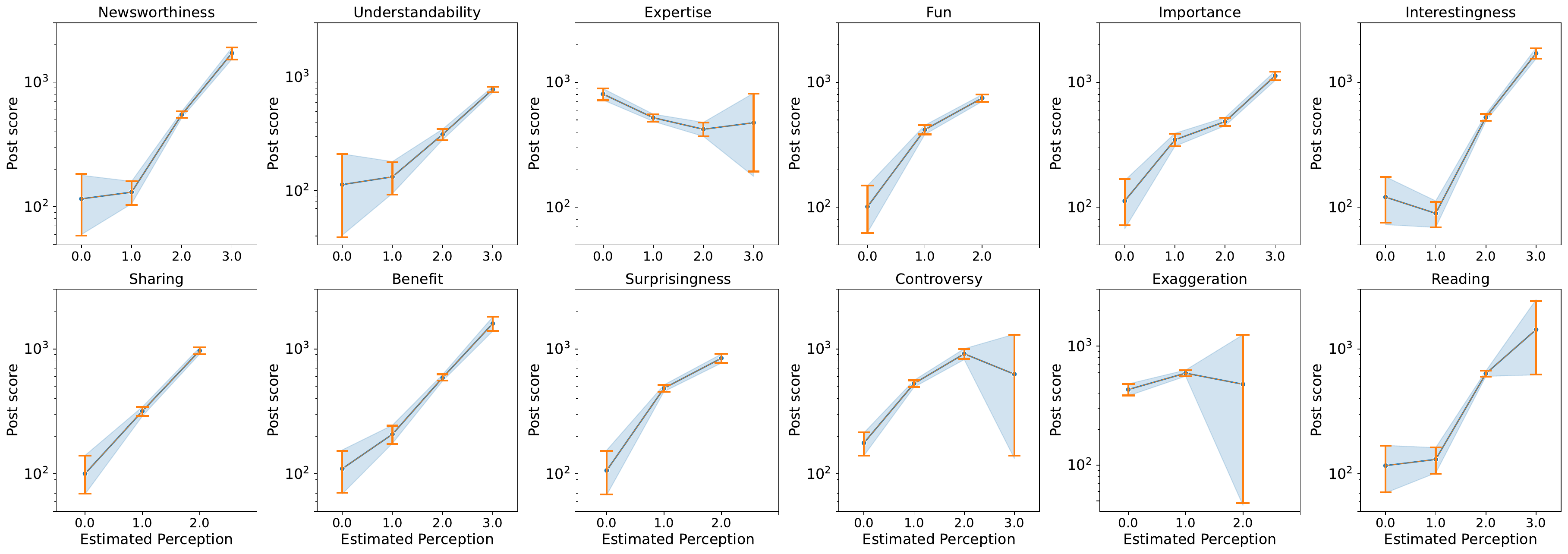}
 \caption{Posts with more positive perception values receive higher scores on Reddit.}
\label{fig:reddit_score_perception_pointplot}
\end{figure*}

\begin{figure*}[t]
 \centering
  \includegraphics[width=\textwidth]{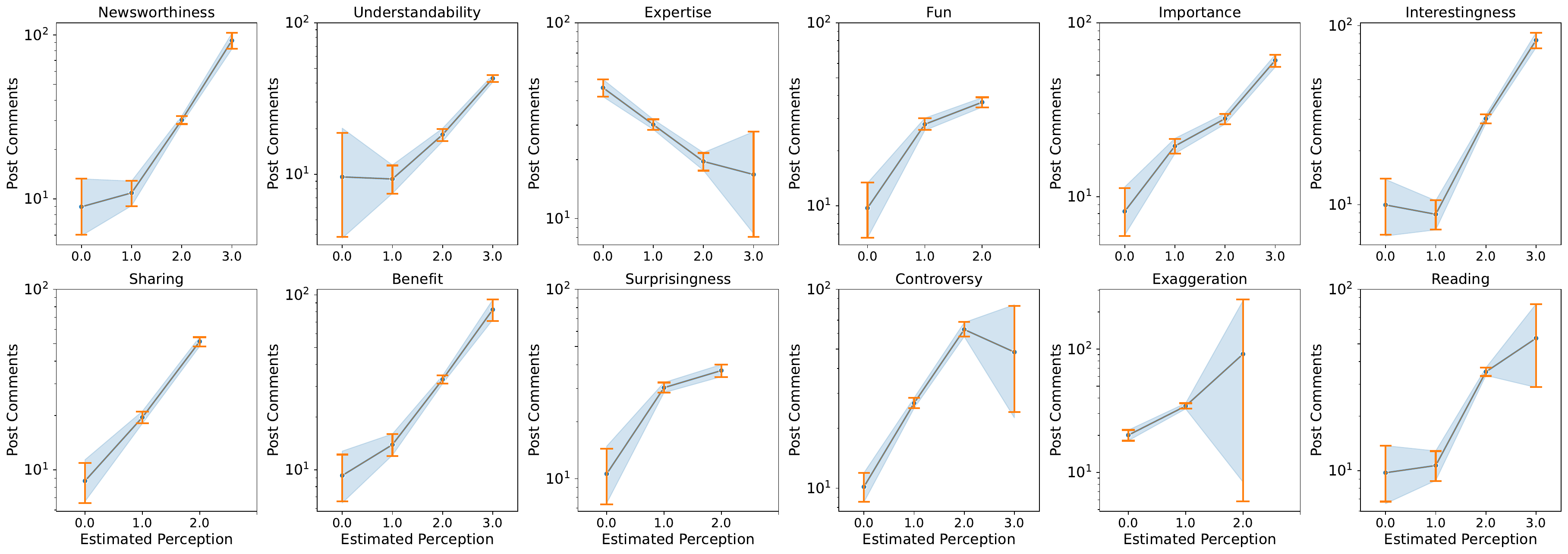}
 \caption{Posts with more positive perception values receive more comments on Reddit.}
\label{fig:reddit_comment_perception_pointplot}
\end{figure*}

\subsection{Predicting and Promoting Public Engagement with Science}

One of the fundamental goals of science communication is to promote public engagement with scientific information \citep{stilgoe2014should}. Social media platforms offer unique opportunities for the public to engage with scientific information. However, predicting public coverage and engagement with science has been notoriously hard \citep{maclaughlin2018predicting} because many types of information are competing for the public's attention in the current media environment \citep{scheufele2013communicating}. Here, we study the connection between estimated perceptions and public engagement with scientific information on Reddit, a popular social media platform that contains many active science-sharing communities.

To study the connections between estimated perceptions and public engagement with social media posts about science, we construct a dataset with 95,465 Reddit posts mentioning 29,099 science-related content r/science subreddit. We conduct two analyses to understand the connection between the perceived values and engagements: (1) a descriptive analysis showing the overall trends across different sciences and (2) a natural experiment where the same science is framed differently in different posts.

\begin{figure}[t!]
  \centering
  \begin{minipage}[t]{0.45\linewidth}  % <-- change [b] to [t]
    \centering
    \includegraphics[width=\linewidth]{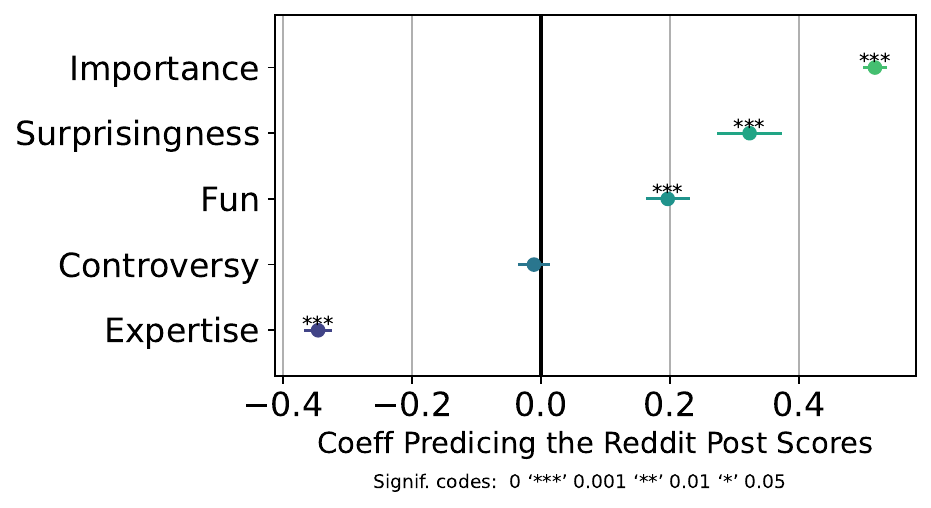}
    \caption{When sharing the same science news story, posts that are framed to be more important, surprising, and fun receive higher scores on Reddit. Posts that require more specialized knowledge to understand receive lower scores.}
    \label{fig:reg_reddit_post_score}
  \end{minipage}%
  \hfill
  \begin{minipage}[t]{0.45\linewidth}  % <-- change [b] to [t]
    \centering
    \includegraphics[width=\linewidth]{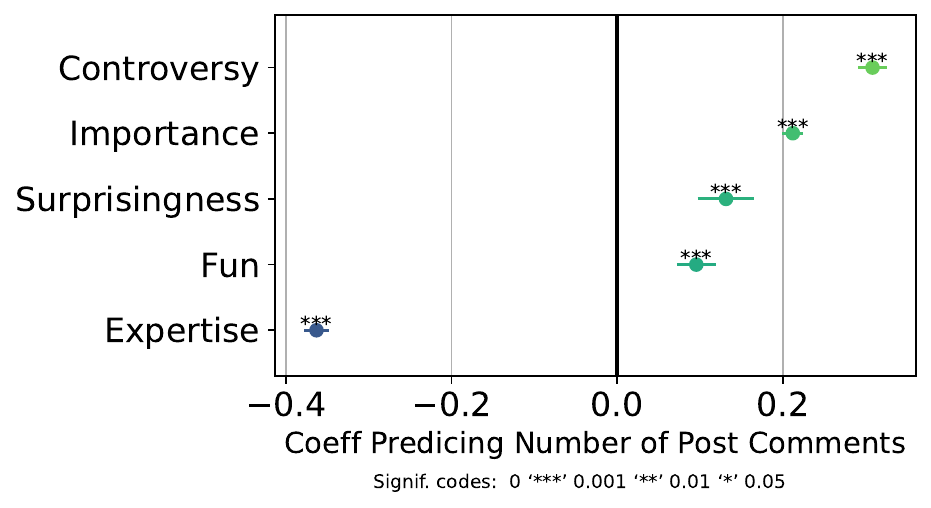}
    \caption{When sharing the same science news story, posts with higher estimated importance, surprisingness, fun, and controversy receive more comments on Reddit. Posts that require more specialized knowledge to understand receive fewer comments.}
    \label{fig:reg_reddit_post_comments}
  \end{minipage}
  %\caption{Main caption for the figure}
  \label{fig:main}
\end{figure}

\paragraph{Posts with more positive perception scores receive more scores and comments}

Figure \ref{fig:reddit_score_perception_pointplot} 
and Figure \ref{fig:reddit_comment_perception_pointplot} 
shows the overall trend of estimated perceptions and public engagement patterns on Reddit. In general, public engagement with science-related posts is positively correlated with the estimated perceptions on most of the dimensions. For example, posts with higher estimated newsworthiness are significantly more likely to receive more upvotes and comments. More specifically, one unit increase of perceived importance could lead to a 68\% increase of post scores. On the contrary, posts that require specialized knowledge to understand receive lower scores and fewer comments. Such a result suggests that the estimated perception values could be used to predict public engagement with science-related posts on social media. 
 
\paragraph{When describing the same science, posts with more positive perceptions attract more comments and upvotes}
Figure \ref{fig:reg_reddit_post_score} and Figure \ref{fig:reg_reddit_post_comments} show the results of the natural experiment. We find that posts with higher \textsc{importance}, \textsc{surprisingness} and \textsc{fun} are more likely to receive higher scores and more comments, reflecting the public's preferences for the presentations of the scientific information. On the contrary, posts that require specialized knowledge to understand receive lower scores and fewer comments, suggesting that science jargon or specialized topics can reduce public engagement with science.  For controversy, we observe different effects for comments and scores: posts with higher controversy values receive significantly more comments, but not scores. Such a result suggests that controversial content attracts more engagement through comments. While the number of comments is usually highly correlated with the post scores, in this setting, it is possible that controversial posts also receive more downvotes, which removes the effect of increased engagement. 
Our result also relates to the literature on framing \citep{pan1993framing, scheufele1999framing}. Existing research has found that framing the same issue in different ways could lead to different communication outcomes and audience perceptions \citep{d2008power}. While we do not study the linguistic features that contribute to different potential perceptions, our model could allow future researchers or journalists to directly estimate the potential public perceptions of the same information, but with different frames.

\section{Discussions}
Effectively engaging the public with science is important but challenging, partly because of the gap between the audience and the science communicators -- science communicators may not be able to fully grasp what the audience is interested in and how they might perceive the scientific information they are receiving. To address this problem, in this study, we design a computational framework that helps to understand and predict public perception of scientific information in the media. Through a series of analyses, we find that the science news consumption frequency is the key factor that drives people's perception of scientific information: people who consume scientific information frequently are more likely to rate science more positively. Such a result reflects a key challenge in science communication that only people who trust and consume science are more likely to positively react to scientific information, but attracting people who do not frequently consume science is challenging. However, our follow-up analysis on Reddit shows a potential hope: even for the same science, framing it in a different way could change the final engagement pattern. Therefore, our computational tool for perception estimation provides science communicators a valuable approach to ``preview'' the potential public perception when drafting the scientific information, helping them craft messages that better engage diverse audiences.

\section{Methods}
To enable large-scale analysis of public perceptions of science, we develop the following computational pipeline combining large-scale crowdsourcing and NLP modeling. 

\paragraph{Raw dataset}
Our raw dataset is obtained from \url{altmetric.com}, which contains 2.4 million media mentions of research papers. We follow the data processing pipeline from existing research \citep{wright2022modeling, pei2021measuring} and further extract a subsample of 128,942 news/article pairs where we are able to collect the full text of the news articles. Our data spans 273 different news outlets and 57,807 different scientific articles in over 100 scientific domains. Appendix \ref{supp:dataset} details more information about the dataset.

\paragraph{Crowdsourcing annotation} We recruit participants from \url{Prolific.com} as it allows us to collect annotations from representative samples for the US and UK population. More specifically, we recruited 1,057 participants from the US and 1,044 participants from the UK. Both are drawn from a representative sample balanced for age, sex, and race. We require each participant to be fluent in English to participate in our study, as the science news articles are written in English. For each news article, we set out to collect 4 labels from the US and the UK samples separately.  We use POTATO to set up all the annotation interface \citep{pei2022potato}. Appendix \ref{supp:dataset} presents all the details of our annotation experiment.

\paragraph{NLP modeling} To allow automatic modeling of perceived news values for each news story, we fine-tune a RoBERTa-Large model \citep{liu2019roberta} on the annotated dataset.  We split the annotated data into train, validation, and test sets following the ratio of 7:1:2. We follow a multi-task regression setting and on the output layer, the model predicts a score ranging from 1-5 for each statement so that for each science news article, we could directly obtain the predictions for all the dimensions. Appendix \ref{supp:model} details the modeling process and results.

\paragraph{Reddit science posting dataset} We first collect all the posts from 2017 to 2021 shared in r/science, a subreddit where users actively share science news stories. We then extract all the shared links in these posts and further extract all the posts on Reddit mentioning the links. We further removed all the duplicated posts and URLs that are only shared once. The final dataset contains 95,465 Reddit posts mentioning 29,099 science-related content. 

\paragraph{Regression} We construct mixed effect models using the estimated perceptions to predict two engagement metrics: the number of comments and the post scores, both log-transformed. Post scores are calculated by subtracting the number of downvotes from the total upvotes. Additionally, comments about scientific information represent an important engagement with scientific information as they mark open discussions about science.
We control the URL as a random effect and also control the subreddit, the domain of the URL, and whether a URL was shared in a subreddit for the first time as a fixed effect. Furthermore, given that public perception correlates with each other, we followed the step-wise regression procedure to gradually remove the variables with the highest variance inflation factor score. The final regression contains five perception dimensions (i.e. \textsc{importance}, \textsc{surprisingness}, \textsc{fun}, \textsc{controversy} and \textsc{expertise}) and other control variables.

\bibliographystyle{unsrt}
\bibliography{custom_cleaned}

\newcommand{\noop}[1]{}
\begin{thebibliography}{10}

\bibitem{fischhoff2013sciences}
Baruch Fischhoff.
\newblock The sciences of science communication.
\newblock {\em Proceedings of the National Academy of Sciences}, 110(supplement\_3):14033--14039, 2013.

\bibitem{bell2003understandings}
Randy~L Bell and Norman~G Lederman.
\newblock Understandings of the nature of science and decision making on science and technology based issues.
\newblock {\em Science education}, 87(3):352--377, 2003.

\bibitem{durant1989public}
John~R Durant, Geoffrey~A Evans, and Geoffrey~P Thomas.
\newblock The public understanding of science.
\newblock {\em Nature}, 340(6228):11--14, 1989.

\bibitem{stilgoe2014should}
Jack Stilgoe, Simon~J Lock, and James Wilsdon.
\newblock Why should we promote public engagement with science?
\newblock {\em Public understanding of science}, 23(1):4--15, 2014.

\bibitem{cheng2008communicating}
Donghong Cheng, Michel Claessens, Toss Gascoigne, Jenni Metcalfe, Bernard Schiele, and Shunke Shi.
\newblock Communicating science in social contexts.
\newblock {\em New Models}, 2008.

\bibitem{scheufele2013communicating}
Dietram~A Scheufele.
\newblock Communicating science in social settings.
\newblock {\em Proceedings of the National Academy of Sciences}, 110(supplement\_3):14040--14047, 2013.

\bibitem{schafer2018different}
Mike~S Sch{\"a}fer, Tobias F{\"u}chslin, Julia Metag, Silje Kristiansen, and Adrian Rauchfleisch.
\newblock The different audiences of science communication: A segmentation analysis of the swiss population’s perceptions of science and their information and media use patterns.
\newblock {\em Public understanding of science}, 27(7):836--856, 2018.

\bibitem{sturgis2004science}
Patrick Sturgis and Nick Allum.
\newblock Science in society: Re-evaluating the deficit model of public attitudes.
\newblock {\em Public understanding of science}, 13(1):55--74, 2004.

\bibitem{atkin1983journalists}
Charles~K Atkin, Judee~K Burgoon, and Michael Burgoon.
\newblock How journalists perceive the reading audience.
\newblock {\em Newspaper research journal}, 4(2):51--63, 1983.

\bibitem{goidel2006exploring}
Kirby Goidel and Matthew Nisbet.
\newblock Exploring the roots of public participation in the controversy over embryonic stem cell research and cloning.
\newblock {\em Political Behavior}, 28:175--192, 2006.

\bibitem{nisbet2009s}
Matthew~C Nisbet and Dietram~A Scheufele.
\newblock What's next for science communication? promising directions and lingering distractions.
\newblock {\em American journal of botany}, 96(10):1767--1778, 2009.

\bibitem{weinburgh1995gender}
Molly Weinburgh.
\newblock Gender differences in student attitudes toward science: A meta-analysis of the literature from 1970 to 1991.
\newblock {\em Journal of Research in science Teaching}, 32(4):387--398, 1995.

\bibitem{achterberg2017science}
Peter Achterberg, Willem De~Koster, and Jeroen Van~der Waal.
\newblock A science confidence gap: Education, trust in scientific methods, and trust in scientific institutions in the united states, 2014.
\newblock {\em Public Understanding of Science}, 26(6):704--720, 2017.

\bibitem{mccright2013influence}
Aaron~M McCright, Katherine Dentzman, Meghan Charters, and Thomas Dietz.
\newblock The influence of political ideology on trust in science.
\newblock {\em Environmental Research Letters}, 8(4):044029, 2013.

\bibitem{olofsson2006attitudes}
Anna Olofsson, Susanna {\"O}hman, and Saman Rashid.
\newblock Attitudes to gene technology: the significance of trust in institutions.
\newblock {\em European Societies}, 8(4):601--624, 2006.

\bibitem{shi2015public}
Jing Shi, Vivianne~HM Visschers, and Michael Siegrist.
\newblock Public perception of climate change: The importance of knowledge and cultural worldviews.
\newblock {\em Risk Analysis}, 35(12):2183--2201, 2015.

\bibitem{feldman2021upping}
Lauren Feldman and P~Sol Hart.
\newblock Upping the ante? the effects of “emergency” and “crisis” framing in climate change news.
\newblock {\em Climatic Change}, 169(1-2):10, 2021.

\bibitem{yeo2015selecting}
Sara~K Yeo, Michael~A Xenos, Dominique Brossard, and Dietram~A Scheufele.
\newblock Selecting our own science: How communication contexts and individual traits shape information seeking.
\newblock {\em The ANNALS of the American Academy of Political and Social Science}, 658(1):172--191, 2015.

\bibitem{hart2012boomerang}
P~Sol Hart and Erik~C Nisbet.
\newblock Boomerang effects in science communication: How motivated reasoning and identity cues amplify opinion polarization about climate mitigation policies.
\newblock {\em Communication research}, 39(6):701--723, 2012.

\bibitem{longnecker2023good}
Nancy Longnecker.
\newblock Good science communication considers the audience.
\newblock In {\em Teaching science students to communicate: a practical guide}, pages 21--30. Springer, 2023.

\bibitem{harcup2017news}
Tony Harcup and Deirdre O’neill.
\newblock What is news? news values revisited (again).
\newblock {\em Journalism studies}, 18(12):1470--1488, 2017.

\bibitem{badenschier2011issue}
Franziska Badenschier and Holger Wormer.
\newblock Issue selection in science journalism: Towards a special theory of news values for science news?
\newblock In {\em The sciences’ media connection--public communication and its repercussions}, pages 59--85. Springer, 2011.

\bibitem{nishal2022crowd}
Sachita Nishal and Nicholas Diakopoulos.
\newblock From crowd ratings to predictive models of newsworthiness to support science journalism.
\newblock {\em Proceedings of the ACM on Human-Computer Interaction}, 6(CSCW2):1--28, 2022.

\bibitem{lee2014newsworthy}
Angela~M Lee and Hsiang~Iris Chyi.
\newblock When newsworthy is not noteworthy: Examining the value of news from the audience's perspective.
\newblock {\em Journalism studies}, 15(6):807--820, 2014.

\bibitem{fischhoff2012communicating}
Baruch Fischhoff.
\newblock Communicating uncertainty fulfilling the duty to inform.
\newblock {\em Issues in Science and Technology}, 28(4):63--70, 2012.

\bibitem{varner2014scientific}
Johanna Varner.
\newblock Scientific outreach: toward effective public engagement with biological science.
\newblock {\em BioScience}, 64(4):333--340, 2014.

\bibitem{morosoli2024scientific}
Jos{\'e}~J Morosoli, Luc{\'\i}a Colodro-Conde, Fiona~Kate Barlow, and Sarah~E Medland.
\newblock Scientific clickbait: Examining media coverage and readability in genome-wide association research.
\newblock {\em Plos one}, 19(1):e0296323, 2024.

\bibitem{hirst2003scientific}
Russel Hirst.
\newblock Scientific jargon, good and bad.
\newblock {\em Journal of technical writing and communication}, 33(3):201--229, 2003.

\bibitem{sharon2014measuring}
Aviv~J Sharon and Ayelet Baram-Tsabari.
\newblock Measuring mumbo jumbo: A preliminary quantification of the use of jargon in science communication.
\newblock {\em Public Understanding of Science}, 23(5):528--546, 2014.

\bibitem{burns1995newspaper}
Risa~B Burns, Mark~A Moskowitz, Michael~A Osband, and Lewis~E Kazis.
\newblock Newspaper reporting of the medical literature.
\newblock {\em Journal of general internal medicine}, 10:19--24, 1995.

\bibitem{galtung1965structure}
Johan Galtung and Mari~Holmboe Ruge.
\newblock The structure of foreign news: The presentation of the congo, cuba and cyprus crises in four norwegian newspapers.
\newblock {\em Journal of peace research}, 2(1):64--90, 1965.

\bibitem{htoo2023medical}
Tint Hla~Hla Htoo, Na~Jin-Cheon, and Michael Thelwall.
\newblock Why are medical research articles tweeted? the news value perspective.
\newblock {\em Scientometrics}, 128(1):207--226, 2023.

\bibitem{sarewitz2004science}
Daniel Sarewitz.
\newblock How science makes environmental controversies worse.
\newblock {\em Environmental science \& policy}, 7(5):385--403, 2004.

\bibitem{friedman1999communicating}
Sharon~M Friedman, Sharon Dunwoody, and Carol~L Rogers.
\newblock {\em Communicating uncertainty: Media coverage of new and controversial science}.
\newblock Routledge, 1999.

\bibitem{stewart2008media}
Craig~O Stewart.
\newblock How a media controversy can influence a scientific publication.
\newblock {\em Discourse Approaches to Politics, Society and Culture (DAPSAC)}, 2008.

\bibitem{caulfield2012science}
Timothy Caulfield and Celeste Condit.
\newblock Science and the sources of hype.
\newblock {\em Public Health Genomics}, 15(3-4):209--217, 2012.

\bibitem{intemann2022understanding}
Kristen Intemann.
\newblock Understanding the problem of “hype”: Exaggeration, values, and trust in science.
\newblock {\em Canadian Journal of Philosophy}, 52(3):279--294, 2022.

\bibitem{neureiter2021trust}
Ariadne Neureiter, Marlis Stubenvoll, Ruta Kaskeleviciute, and J{\"o}rg Matthes.
\newblock Trust in science, perceived media exaggeration about covid-19, and social distancing behavior.
\newblock {\em Frontiers in Public Health}, 9:670485, 2021.

\bibitem{yin2022public}
Yian Yin, Yuxiao Dong, Kuansan Wang, Dashun Wang, and Benjamin~F Jones.
\newblock Public use and public funding of science.
\newblock {\em Nature human behaviour}, 6(10):1344--1350, 2022.

\bibitem{shearer2019americans}
Elisa Shearer and Elizabeth Grieco.
\newblock Americans are wary of the role social media sites play in delivering the news.
\newblock {\em Pew Research Center}, 2, 2019.

\bibitem{bednarek2016investigating}
Monika Bednarek.
\newblock Investigating evaluation and news values in news items that are shared through social media.
\newblock {\em Corpora}, 11(2):227--257, 2016.

\bibitem{national2017communicating}
{National Academies of Sciences, Engineering, and Medicine}.
\newblock Communicating science effectively: A research agenda.
\newblock 2017.

\bibitem{bak2001education}
Hee-Je Bak.
\newblock Education and public attitudes toward science: Implications for the “deficit model” of education and support for science and technology.
\newblock {\em Social Science Quarterly}, 82(4):779--795, 2001.

\bibitem{osborne2003attitudes}
Jonathan Osborne, Shirley Simon, and Sue Collins.
\newblock Attitudes towards science: A review of the literature and its implications.
\newblock {\em International journal of science education}, 25(9):1049--1079, 2003.

\bibitem{jones2000gender}
M~Gail Jones, Ann Howe, and Melissa~J Rua.
\newblock Gender differences in students' experiences, interests, and attitudes toward science and scientists.
\newblock {\em Science education}, 84(2):180--192, 2000.

\bibitem{blank2015does}
Joshua~M Blank and Daron Shaw.
\newblock Does partisanship shape attitudes toward science and public policy? the case for ideology and religion.
\newblock {\em The ANNALS of the American Academy of Political and Social Science}, 658(1):18--35, 2015.

\bibitem{maclaughlin2018predicting}
Ansel MacLaughlin, John Wihbey, and David Smith.
\newblock Predicting news coverage of scientific articles.
\newblock In {\em Proceedings of the International AAAI Conference on Web and Social Media}, volume~12, 2018.

\bibitem{pan1993framing}
Zhongdang Pan and Gerald~M Kosicki.
\newblock Framing analysis: An approach to news discourse.
\newblock {\em Political communication}, 10(1):55--75, 1993.

\bibitem{scheufele1999framing}
Dietram~A Scheufele.
\newblock Framing as a theory of media effects.
\newblock {\em Journal of communication}, 49(1):103--122, 1999.

\bibitem{d2008power}
Paul D'Angelo and Matthew Lombard.
\newblock The power of the press: The effects of press frames in political campaign news on media perceptions.
\newblock {\em Atlantic Journal of Communication}, 16(1):1--32, 2008.

\bibitem{wright2022modeling}
Dustin Wright, Jiaxin Pei, David Jurgens, and Isabelle Augenstein.
\newblock Modeling information change in science communication with semantically matched paraphrases.
\newblock {\em arXiv preprint arXiv:2210.13001}, 2022.

\bibitem{pei2021measuring}
Jiaxin Pei and David Jurgens.
\newblock Measuring sentence-level and aspect-level (un) certainty in science communications.
\newblock {\em arXiv preprint arXiv:2109.14776}, 2021.

\bibitem{pei2022potato}
Jiaxin Pei, Aparna Ananthasubramaniam, Xingyao Wang, Naitian Zhou, Apostolos Dedeloudis, Jackson Sargent, and David Jurgens.
\newblock Potato: The portable text annotation tool.
\newblock In {\em Proceedings of the 2022 Conference on Empirical Methods in Natural Language Processing: System Demonstrations}, 2022.

\bibitem{liu2019roberta}
Yinhan Liu, Myle Ott, Naman Goyal, Jingfei Du, Mandar Joshi, Danqi Chen, Omer Levy, Mike Lewis, Luke Zettlemoyer, and Veselin Stoyanov.
\newblock Roberta: A robustly optimized bert pretraining approach.
\newblock {\em arXiv preprint arXiv:1907.11692}, 2019.

\bibitem{caple2018news}
Helen Caple.
\newblock News values and newsworthiness.
\newblock In {\em Oxford research encyclopedia of communication}. 2018.

\bibitem{dalecki2009news}
Linden Dalecki, Dominic~L Lasorsa, and Seth~C Lewis.
\newblock The news readability problem.
\newblock {\em Journalism Practice}, 3(1):1--12, 2009.

\bibitem{agarwal1992analysis}
Tara Agarwal.
\newblock {\em An analysis of the readability of science and technology reporting in four national newspapers}.
\newblock PhD thesis, Loughborough University of Technology, 1992.

\bibitem{bourk2018entertainment}
Michael Bourk, Bienvenido Le{\'o}n, and Lloyd~S Davis.
\newblock Entertainment in science: Useful in small doses.
\newblock In {\em Communicating science and technology through online video}, pages 90--106. Routledge, 2018.

\bibitem{narasimhan2001controversy}
Marehalli~G Narasimhan.
\newblock Controversy in science.
\newblock {\em Journal of biosciences}, 26(3):299--304, 2001.

\bibitem{hayes2007answering}
Andrew~F Hayes and Klaus Krippendorff.
\newblock Answering the call for a standard reliability measure for coding data.
\newblock {\em Communication methods and measures}, 1(1):77--89, 2007.

\bibitem{tavakol2011making}
Mohsen Tavakol and Reg Dennick.
\newblock Making sense of cronbach's alpha.
\newblock {\em International journal of medical education}, 2:53, 2011.

\bibitem{dittrich2007paired}
Regina Dittrich, Brian Francis, Reinhold Hatzinger, and Walter Katzenbeisser.
\newblock A paired comparison approach for the analysis of sets of likert-scale responses.
\newblock {\em Statistical Modelling}, 7(1):3--28, 2007.

\bibitem{reidsma2008reliability}
Dennis Reidsma and Jean Carletta.
\newblock Reliability measurement without limits.
\newblock {\em Computational Linguistics}, 34(3):319--326, 2008.

\bibitem{artstein2008inter}
Ron Artstein and Massimo Poesio.
\newblock Inter-coder agreement for computational linguistics.
\newblock {\em Computational linguistics}, 34(4):555--596, 2008.

\end{thebibliography}

\appendix

\section{Twelve-dimension framework for public perception of science}
\label{supp:framework-table}
Table \ref{tab:news_value_dimensions} shows the statements and descriptions for the twelve-dimensional framework for measuring public perception of science. 

\begin{table}
\centering
\resizebox{0.9\textwidth}{!}{
\begin{tabular}{p{4cm} p{8cm} p{11cm}}
\rowcolor{lightgrey}\hline
Category & Statement & Description \\ \hline

\rowcolor{white}
\textsc{Newsworthiness} & The science news story should be published in the news & In journalism, newsworthiness is considered as the core dimension of news stories and is usually operationalized as “worthy of being published as news” \citep{caple2018news}. Here we measure the perceived newsworthiness from the perspectives of the public. \\

\rowcolor{lightgrey}
\textsc{Understandability} & I can understand the science news story & Understandability measures whether a given science news story can be understood by the public. Existing studies consider understandability as an important dimension for news stories \citep{dalecki2009news, agarwal1992analysis}. \\

\rowcolor{white}
\textsc{Expertise} & Understanding the science news story requires specialized knowledge & Scientific information may involve specialized knowledge and the \textsc{Expertise} dimension measures whether a given news story requires specialized knowledge to understand. \\

\rowcolor{lightgrey}
\textsc{Importance} & The science news story tackles an important issue & Importance has been considered as one of the core value factors of science news stories \citep{badenschier2011issue}. Here we measure how the public perceives the importance of the scientific information presented in a news article. \\

\rowcolor{white}
\textsc{Fun} & The science news story is fun to read & Entertainment has long been considered as one of the core news values in journalism \citep{harcup2017news} and the public also actively perceives the entertainment value of science \citep{bourk2018entertainment}. Here we measure whether a given scientific information is perceived as fun and entertaining to the public. \\

\rowcolor{lightgrey}
\textsc{Surprisingness} & The scientific finding seems surprising to me & Science sometimes provides surprising results that contradict people’s existing beliefs and surprisingness has been considered as an important factor for science news values \citep{badenschier2011issue}. Here we measure whether the public perceives a given science news story as surprising. \\

\rowcolor{white}
\textsc{Controversy} & The scientific finding could be controversial & Science tackles many controversial topics \citep{narasimhan2001controversy} and here we measure whether a given science is perceived as controversial by the public. \\

\rowcolor{lightgrey}
\textsc{Exaggeration} & This science news story is overstated or exaggerated & Exaggeration is a common issue in science reporting \citep{caulfield2012science} and can potentially undermine the public trust in science \citep{intemann2022understanding}. Here, we measure the subjective perceptions of exaggeration in scientific information. \\

% --- Interestingness (2 rows) ---
\rowcolor{white}
\multirow{1}{4cm}{\textsc{Interestingness}} 
  & The science news story sounds interesting to me  
  & \multirow{2}{11cm}{Perceived interestingness is considered as a key factor affecting people’s perception of the overall newsworthiness \citep{lee2014newsworthy}. Here we measure the perceived interestingness from two dimensions: (1) the perspective from an individual and (2) the perceived interestingness for the general public.} \\
\rowcolor{white}
  & The science news story could be interesting to the general public & \\

% --- Benefit (7 rows) ---
\rowcolor{lightgrey} 
  & Who will benefit from learning about this finding? & \multirow{7}{10cm}{Benefit is one of the core dimensions of science news values \citep{badenschier2011issue}. Here, we measure the perceived benefits of a given science news story from multiple latent dimensions.} \\
\rowcolor{lightgrey} & The general public & \\
\rowcolor{lightgrey} & A segment of the public & \\
\rowcolor{lightgrey} \multirow{1}{4cm}{\textsc{Benefit}} & Policy makers & \\
\rowcolor{lightgrey} & Companies in the related industries & \\
\rowcolor{lightgrey} & I learned something useful from the science news story & \\
\rowcolor{lightgrey} & Knowing about this science could benefit a lot of people & \\

% --- Sharing (3 rows) ---
\rowcolor{white}

  & I would share this science news story with someone I know directly & \multirow{3}{10cm}{The sharability of a news article has been considered as an important dimension of the news values \citep{harcup2017news}. Here we measure the public’s sharing willingness in different imagined settings (e.g.\ forums and friends).} \\
\rowcolor{white} \multirow{1}{4cm}{\textsc{Sharing}}
  & I would share this science news story with a wider forum like a mailing list, Twitter, Reddit & \\
\rowcolor{white}
  & I would be unlikely to share this science news story with anyone & \\

% --- Reading (6 rows) ---
\rowcolor{lightgrey}

  & If I’m browsing news articles from \_\_, I would be likely to open and read the science news story above & \multirow{6}{10cm}{Reading willingness measures the audience’s willingness to read the presented information in different venues (e.g.\ general news outlets, science and technology media and other popular media). Modeling this dimension could further help to understand the public’s interest in reading about science news stories.} \\
\rowcolor{lightgrey} & General news outlets (e.g.\ BBC, New York Times, Fox News) & \\
\rowcolor{lightgrey} \multirow{1}{4cm}{\textsc{Reading}} & Science and Technology media (e.g.\ Scientific American, National Geographic) & \\
\rowcolor{lightgrey} & Other popular printing media (e.g.\ Vogue, GQ, Elle) & \\
\rowcolor{lightgrey} & Other popular media (e.g.\ TV and radio) & \\
\rowcolor{lightgrey} & It should not be published in public media outside the science community & \\ \hline

\end{tabular}
}
\caption{The twelve-dimension framework to measure public perceptions of scientific information in media.}
\label{tab:news_value_dimensions}
\end{table}

\section{Building a New Dataset to Model Public Perception of Scientific Information}
\label{supp:dataset}

Data resources are key to scientific research. 
To support long-term research on public perception of scientific information and promote public engagement with science, we create a new large-scale dataset containing nuanced ratings for individual science news stories as well as detailed background information of the participants. In this section, I describe the creation process of this dataset.

\subsection{Raw Dataset}
Our raw dataset is obtained from \url{altmetric.com} which contains 2.4 million media mentions of research papers. We follow the processing pipeline from previous research \citep{wright2022modeling, pei2021measuring} and further extract a subsample of 128,942 news/article pairs where we are able to collect the full text of the news articles. Our data spans 273 different news outlets and 57,807 different scientific articles in over 100 scientific domains.

\subsection{Crowdsourcing with Representative Samples}
To measure the public perception of science news stories, we designed the following pipeline to create a large-scale crowdsourcing annotation task.

\paragraph{Data Cleaning}
For our annotation task, we aim to present the full news articles while excluding information that might bias the annotators' perception of scientific information. To this end, we designed a data-cleaning pipeline that removes the following types of information: (1) Outlet: people's perception of news articles might be biased by their pre-existing attitudes towards the outlet (2) Other meta information like author, date, city, and URL, as such information might create an impression that the presented news article has already been published, which may further bias the ratings. %Further more, we focus on science news stories published from 2011 to 2019. 

\paragraph{Data Sampling for Annotation}
The overall goal of our data sampling is to provide a generally balanced sample for annotation across different settings. We design a sampling procedure that balances the type of outlet, the science domain, and the popularity of the paper. Below we describe our sampling process for a batch of around 1000 news articles:

\begin{enumerate}
    \item Sampling based on coverage types. We first categorize each outlet into three categories: General, Press Release, and SciTech blogs. Furthermore, we balancedly sample the same amount of papers (e.g. 80 in this case) from three settings: (1) mentioned in all three types of outlets (2) mentioned in both PR and SciTech outlets (3) mentioned in both PR and General outlets. For each selected paper, we randomly sample 1 article from the corresponding category. Such a process will lead to 560 sampled news articles. 
    \item Sampling based on outlet types. We further expand our sample by randomly sampling 50 news articles from each type of outlet separately. This step leads to 810 news stories.
    \item Upsampling based on domains. To balance the stories in different science domains, we further upsample another 50 news stories in Social Science, 50 in Humanities, and 100 in engineering, which leads to 1,010 stories.
    \item Upsampling based on popularity. Previous steps may lead to a sample of papers with relatively low popularity, which might have lower news values than papers that are covered in many news stories. In this step, we further upsample another 30 news articles that cover popular research articles (coverage count larger than 30).  
\end{enumerate}

After this process, we are able to obtain a sample of news stories that are balanced for the type of coverage, type of outlet, domain, and popularity of the paper.

\paragraph{Particpants} We recruit participants from \url{Prolific.com} as it allows us to collect annotations from representative samples for the US and UK population. More specifically, we recruit 1,057 participants from the US and 1,044 participants from the UK. Both are drawn from a representative sample balanced for age, sex, and race. We require each participant to be fluent in English to participate in our study as the science news articles are written in English. For each news article, we set to collect 4 labels from the US and the UK sample separately. 

\begin{table}[tbp]
  \centering
  \resizebox{.99\textwidth}{!}{%
  \begin{tabular}{llcc}
    \hline
    \rowcolor{gray!30}
    Category & \textbf{Statement} & \textbf{K's $\alpha$} & \textbf{C's $\alpha$} \\
    \hline
    \textsc{Newsworthiness} & Should be published in news                                             & 0.130 & {}     \\ 
    \rowcolor{lightgrey}
    \textsc{Importance}     & Tackles an important issue                                              & 0.311 & {}     \\
    \textsc{Expertise}      & Needs specialized knowledge                                              & 0.248 & {}     \\ 
    \rowcolor{lightgrey}
    \textsc{Understandability} & I understand the news story                                            & 0.185 & {}     \\
    \textsc{Fun}            & News story is fun to read                                                & 0.122 & {}     \\ 
    \rowcolor{lightgrey}
    \textsc{Controversy}     & Finding could be controversial                                           & 0.147 & {}     \\
    \textsc{Surprisingness}  & Finding is surprising to me                                               & 0.096 & {}     \\ 
    \rowcolor{lightgrey}
    \textsc{Exaggeration}    & This science news story is overstated or exaggerated                     & 0.045 & {}     \\[1ex]
    \textsc{Interestingness} & Could interest the general public                                        & 0.137 & \multirow{2}{*}{0.77} \\
    \textsc{Interestingness} & The science news story sounds interesting to me                           & 0.092 & {}                        \\[1ex]
    \rowcolor{lightgrey}
    \textsc{Benefit} & Could benefit policy makers                                                    & 0.233 & \multirow{6}{*}{0.84} \\
    \textsc{Benefit} & Could benefit many people                                                      & 0.220 & {}                        \\ 
    \rowcolor{lightgrey}
    \textsc{Benefit} & Could benefit related industry companies                                       & 0.156 & {}                        \\  
    \textsc{Benefit} & Could benefit the general public                                               & 0.166 & {}                        \\ 
    \rowcolor{lightgrey}
    \textsc{Benefit} & Could benefit a segment of the public                                          & 0.113 & {}                        \\ 
    \textsc{Benefit} & I learned something useful                                                      & 0.082 & {}                        \\[1ex]
    \textsc{Sharing} & I would share this science news story with someone I know directly            & 0.050 & \multirow{3}{*}{0.81} \\
    \textsc{Sharing} & I would be unlikely to share this science news story with anyone             & 0.024 & {}                        \\
    \textsc{Sharing} & I would share this science news story with a wider forum (e.g., Twitter)     & 0.022 & {}                        \\[1ex]
    \rowcolor{lightgrey}
    \textsc{Reading} & Would be willing to read it from general news outlets (e.g., BBC, NYT)        & 0.083 & \multirow{6}{*}{0.83} \\
    \textsc{Reading} & Would be willing to read it from other popular media (TV, radio)              & 0.074 & {}                        \\ 
    \rowcolor{lightgrey}
    \textsc{Reading} & Would be willing to read it from other print outlets (Vogue, Time, etc.)     & 0.082 & {}                        \\
    \textsc{Reading} & Would be willing to read it in my social-media feed                           & 0.047 & {}                        \\ 
    \rowcolor{lightgrey}
    \textsc{Reading} & Would be willing to read it from Sci\&Tech outlets (NatGeo, SciAm)            & 0.070 & {}                        \\ 
    \textsc{Reading} & It should not be published outside the science community                     & 0.012 & {}                        \\
    \hline
  \end{tabular}}
  \caption{Krippendorff's $\alpha$ and Cronbach's $\alpha$ for rating the public perceptions of science news stories.}
  \label{tab:iaa-cronbach}
\end{table}

%We do not apply other filters in the hope to collect annotations from a 

\paragraph{Annotation Procedure} We first ask each annotator to read the introduction of our study and then consent to participate in the annotation. Once they agree to participate, we display 5 randomly sampled science news stories in a sequence. To measure the public perception of science news values, we follow the framework introduced in Sec \ref{sec:framework_news_value} and ask each annotator to rate the given science news stories based on the 25 statements in Table \ref{tab:news_value_dimensions}. Each annotator is asked to rate the 25 statements for each science news story following a 1-5 Likert scale, where 1 means strongly disagree and 5 means strongly agree. %Table \ref{tab:news_value_dimensions} provides the detailed statements used in our annotation task. 
After the annotation phase, each annotator is asked to answer a list of questions regarding their demographic information, educational background, experience/interest in science, and political orientation. %Table \bn provides a detailed list of the post-annotation questions. 

\subsection{Quality Check and Post Processing}
Our final dataset contains 10,489 annotations for 1,506 news articles. To evaluate the quality of the created dataset, we calculate Krippendorrf's $\alpha$ and Cronbach's $\alpha$ to evaluate the reliability of the inter-annotator agreement and the internal consistency of the scales. 

\paragraph{Inter-annotator Agreement} Table \ref{tab:iaa-cronbach} shows the inter-annotator agreement as calculated by Krippendorff's $\alpha$ \citep{hayes2007answering} for each dimension. The average IAA across all the statements is 0.11, suggesting that raters' perceptions of science news stories is subjective. Despite its subjectivity, we observe that the IAA for rating certain statements is generally higher, for example, the IAA for ``The finding needs specialized knowledge'', ``Could benefit many people'' and ``Tackles an important issue'' are generally above 0.2, suggesting that the annotators have more agreement when rating relatively more objective dimensions. While the overall IAA seems low in our data, it is worth noting that a similar level of IAA was observed in a similar annotation task conducted by \citet{nishal2022crowd}. \citet{nishal2022crowd} collected newsworthiness ratings from both expert raters and crowd workers. The IAA between expert raters is 0.16 and ranges from 0.09 to 0.18 for crowd workers, suggesting that the low IAA in our task is potentially caused by natural subjectivity instead of low-quality annotations.

\paragraph{Internal Consistency} To evaluate the internal consistency of the instruments used for measuring public perceptions, We calculate Cronbach's $\alpha$ \citep{tavakol2011making} for each group of statements including \textsc{Interestingness}, \textsc{Benefit}, \textsc{Sharing}, and \textsc{Reading} and the C's $\alpha$ is 0.77, 0.84, 0.81, 0.83 respectively, suggesting that the scales used to measure public perceptions are generally reliable.

\paragraph{Post Processing} For dimensions measured with multiple latent statements (e.g. \textsc{Reading} and \textsc{Sharing}), we average all the relevant ratings from an annotator to create a unified score for the dimension. After this step, each news article receives 12 scores from a single annotator. We then average the ratings of all the annotators to obtain the overall perception value. As a result, the final dataset contains perception scores for 1,506 news stories, and each story has 12 scores representing the public perceptions of it on the 12 dimensions. 

\paragraph{Usage for Modeling Training} One of the final goals of our study is to build NLP models that are able to automatically estimate public perceptions of scientific information. Therefore, it is worth discussing the suitability of using this data for building machine learning models. First of all, our post-processing steps ensure that we are modeling the mean relative perceptions instead of the numerical ratings. More specifically, when a news article receives a high average newsworthiness rating, it is more likely to be more newsworthy than another article with a low average score. To quantify the reliability of our final average scores as a relative measurement of news perceptions, we follow a similar procedure as \cite{dittrich2007paired} to convert the Likert ratings into ranking scores. We then calculate the correlation between the average rating and the ranking score on all the dimensions. The average correlation across all the dimensions is 0.8, suggesting that the average rating could reliably measure the relative perception orders across articles.  Secondly, while we observe relatively low inter-annotator agreement, existing studies have suggested that low-IAA data can still be used to train reliable machine learning models \citep{reidsma2008reliability, artstein2008inter}. For example, in a similar task modeling the newsworthiness of research articles \citep{nishal2022crowd}, while the IAAs between crowd workers are around 0.1, a model trained on the annotated data could still precisely predict the expert ratings for newsworthiness. As we are following a similar post-processing pipeline as \citep{nishal2022crowd}, we argue that our annotations, while with relatively low IAA (from 0.01 to 0.3), could still be used to train NLP models for automatically predicting public perceptions of scientific information.

\section{Computational Modeling of Public Perceptions}
\label{supp:model}
To model the public perceptions of scientific information at scale, we develop a data-driven NLP pipeline that is able to estimate the public perception of each science news article. %In this section, we discuss the data collection, data annotation, NLP modeling, and preliminary results.

\paragraph{Model} To allow automatic modeling of perceived news values for each news story, we fine-tune a RoBERTa-Large model \citep{liu2019roberta} on the annotated dataset.  We split the annotated data into train, validation, and test set following the ratio of 7:1:2. We concatenate the title and the body of each news article as the input document.  We follow a multi-task regression setting and on the output layer, the model predicts a score ranging from 1-5 for each statement, so that for each science news article, we could directly obtain the predictions for all the dimensions. We fine-tune each model for 10 epochs and then choose the best-performing model on the validation set as the final model and further evaluate the model's performance on the hold-out test set.

\paragraph{Result} Table \ref{tab:model_performance} shows the model performances on the test set.  We found that the model is generally able to predict the perception of science news stories on most dimensions. The correlations for 7 out of the 12 dimensions are above 0.5. However, the model performs poorly on dimensions that are relatively more subjective. For example, the prediction performances for exaggerating and sharing are only 0.31 on news stories. Overall, our experiment suggests that automatically measuring the public perceptions of scientific information in media is doable but also challenging. 

%\when there is a relatively high agreement. For example, Pearson's $r$ for ``Needs specialized knowledge'' and ``I understand the news story'' are both above 0.7. For dimensions with lower IAA, the model prediction performances decrease accordingly, suggesting that it is relatively difficult to predict the population average when there are significant individual differences. 

\begin{table}[h]
\centering
\resizebox{0.5\textwidth}{!}{%
\begin{tabular}{lcccc}
\hline
\rowcolor{gray!30}
Dimension     & News & Reddit & Twitter & Paper \\ \hline \rowcolor{lightgrey} 
\textsc{Newsworthiness}     & 0.54 & 0.69   & 0.42 & 0.52    \\
\textsc{Reading}            & 0.44 & 0.52   & 0.49 &     \\ \rowcolor{lightgrey}
\textsc{Understandability}  & 0.71 & 0.75   & 0.79 & 0.65    \\
\textsc{Expertise}          & 0.74 & 0.40   & 0.39 & 0.66*    \\ \rowcolor{lightgrey}
\textsc{Fun}                & 0.60 & 0.76   & 0.72 &    \\
\textsc{Importance}         & 0.77 & 0.62   & 0.20 &    \\\rowcolor{lightgrey}
\textsc{Interestingness}    & 0.52 & 0.34   & 0.14  &  \\
\textsc{Sharing}            & 0.31 & 0.69   & 0.48 &   \\\rowcolor{lightgrey}
\textsc{Benefit}            & 0.67 & 0.35   & 0.24 &   \\
\textsc{Surprisingness}     & 0.41 & 0.56   & 0.57  &  \\\rowcolor{lightgrey}
\textsc{Controversy}        & 0.49 & 0.16   & 0.05 & 0.44    \\
\textsc{Exaggeration}       & 0.31 & 0.38   & 0.16 &    \\ \hline \rowcolor{lightgrey}
Overall            & 0.54 & 0.52 & 0.39  &   \\ \hline
\end{tabular}%
}
\caption{Model performances as measured by Pearson's $r$ over different test sets. The model generally performs well on News and Reddit posts but performs poorly on tweets. ``Paper'' refers to the Arxiv abstract newsworthiness dataset \citep{nishal2022crowd}. We calculated the correlation between our model's predictions and crowd ratings for the corresponding category. Overall, our model also shows good performance on the Arxiv abstract newsworthiness dataset. * means that an absolute correlation score is used because the two measured dimensions are conceptually opposite.}
\label{tab:model_performance}
\end{table}

\paragraph{Model Generalizability on Social Media Text} To evaluate the model performance in other domains, we created another test set for Reddit posts and tweets mentioning research articles or news stories. A similar annotation pipeline was adopted to annotate 50 Reddit posts and 50 tweets about scientific information. Each post is annotated by 8 annotators and Table \ref{tab:iaa-reddit-twitter} shows the internal consistency and inter-annotator agreement for the two datasets. We further run the trained model over this dataset and the results are shown in Table \ref{tab:model_performance}. Overall we observe that the model performs well on Reddit posts but performs poorer on tweets. This is potentially because the tweets are usually shorter and may contain more noise.

\begin{table}[h]
\centering
\resizebox{0.6\textwidth}{!}{%
\begin{tabular}{lllll}
\rowcolor{lightgrey}
\hline
& \textbf{K's $\alpha$} & \textbf{K's $\alpha$} & \textbf{C's $\alpha$} & \textbf{C's $\alpha$} \\
\rowcolor{lightgrey}
& Reddit & Twitter & Reddit & Twitter \\
\rowcolor{white}
\hline
Newsworthiness & 0.41 & 0.32 & & \\
\rowcolor{lightgrey}
Understandability & 0.36 & 0.45 & & \\
\rowcolor{white}
Expertise & 0.33 & 0.53 & & \\
\rowcolor{lightgrey}
Fun & 0.31 & 0.06 & & \\
\rowcolor{white}
Importance & 0.60 & 0.46 & & \\
\rowcolor{lightgrey}
Surprisingness & 0.26 & 0.13 & & \\
\rowcolor{white}
Controversy & 0.31 & 0.28 & & \\
\rowcolor{lightgrey}
Exaggeration & 0.08 & 0.23 & & \\
\rowcolor{white}
Interestingness & 0.40 & 0.33 & 0.79 & 0.75 \\
\rowcolor{lightgrey}
Benefit & 0.43 & 0.41 & 0.86 & 0.84 \\
\rowcolor{white}
Sharing & 0.15 & 0.23 & 0.73 & 0.81 \\
\rowcolor{lightgrey}
Reading & 0.30 & 0.24 & 0.81 & 0.77 \\
\hline
\end{tabular}%
}
\caption{Krippendorff's $\alpha$ and Cronbach's $\alpha$ for rating the public perceptions of Reddit and Twitter posts. We observe similar or higher levels of reliability and internal consistency compared with news annotations.}
\label{tab:iaa-reddit-twitter}
\end{table}

\paragraph{Model Generalizability on Paper Abstracts} To test the generalizability of our model on research articles, we also evaluate model performance on the Arxiv newsworthiness dataset \citep{nishal2022crowd}, which contains readability, newsworthiness, and controversy labels for computer science paper abstracts. We calculate the correlation between our model predictions and the ground truth crowd labels for overlapped dimensions. More specifically, for \textsc{newsworthiness} and \textsc{controversy}, we directly calculate the correlations. For \textsc{Understandability} and \textsc{Expertise}, we calculate the correlation with the readability labels in the Arxiv newsworthiness dataset. As shown in Table \ref{tab:model_performance}, we found moderate to high correlations between our model predictions and the ground truth labels. For example, the Pearson's $r$ for \textsc{newsworthiness} is 0.52 and is 0.65 for \textsc{understandability}. As a comparison, the correlation between the ground truth expert ratings is 0.33. Since the Arxiv abstracts were all sampled from computer science papers, such a result suggests that our model is able to capture nuanced differences for content in the same science domain despite the low IAA of the annotations. 

%\paragraph{Generalizability test on research articles} 

\end{document}